\documentclass[lettersize,journal]{IEEEtran}
\usepackage{amsmath,amsfonts}
\usepackage{algorithmic}
\usepackage{algorithm}
\usepackage{array}
\usepackage[caption=false,font=normalsize,labelfont=sf,textfont=sf]{subfig}
\usepackage{url}
\usepackage{graphicx}
\usepackage[table,xcdraw]{xcolor}
\usepackage{scalerel}
\usepackage{tikz}
\usetikzlibrary{svg.path}

\definecolor{orcidlogocol}{HTML}{A6CE39}
\tikzset{
  orcidlogo/.pic={
    \fill[orcidlogocol] svg{M256,128c0,70.7-57.3,128-128,128C57.3,256,0,198.7,0,128C0,57.3,57.3,0,128,0C198.7,0,256,57.3,256,128z};
    \fill[white] svg{M86.3,186.2H70.9V79.1h15.4v48.4V186.2z}
                 svg{M108.9,79.1h41.6c39.6,0,57,28.3,57,53.6c0,27.5-21.5,53.6-56.8,53.6h-41.8V79.1z M124.3,172.4h24.5c34.9,0,42.9-26.5,42.9-39.7c0-21.5-13.7-39.7-43.7-39.7h-23.7V172.4z}
                 svg{M88.7,56.8c0,5.5-4.5,10.1-10.1,10.1c-5.6,0-10.1-4.6-10.1-10.1c0-5.6,4.5-10.1,10.1-10.1C84.2,46.7,88.7,51.3,88.7,56.8z};
  }
}
\newcommand\orcidicon[1]{\href{https://orcid.org/#1}{\mbox{\scalerel*{
\begin{tikzpicture}[yscale=-1,transform shape]
\pic{orcidlogo};
\end{tikzpicture}
}{|}}}}
\usepackage{booktabs}
\usepackage[breaklinks,colorlinks]{hyperref}
\newcommand{\settablefont}{\fontsize{7.5}{13.0}\selectfont}
\usepackage{orcidlink}
\usepackage{multirow}
\usepackage{physics}
\usepackage{mathtools}
\usepackage{amsmath}
\usepackage{ulem}
\usepackage{float}
\usepackage{stfloats}
\usepackage{caption}
\usepackage{colortbl}
\usepackage[table]{xcolor}
\usepackage{makecell}
\usepackage[pagewise]{lineno}
\newcolumntype{M}[1]{>{\centering\arraybackslash}m{#1}}

\begin{document}

\title{
Fully Exploiting Vision Foundation Model's Profound Prior Knowledge for Generalizable\\RGB-Depth Driving Scene Parsing
} 

\author{Sicen Guo$^{\orcidicon{0009-0000-8079-8056}\,}$, Tianyou Wen$^{\orcidicon{0009-0000-6172-510X}\,}$, Chuang-Wei Liu$^{\orcidicon{0000-0003-0260-6236}\,}$, \\ Qijun Chen$^{\orcidicon{0000-0001-5644-1188}\,}$~\IEEEmembership{Senior Member,~IEEE}, Rui Fan$^{\orcidicon{0000-0003-2593-6596}\,}$~\IEEEmembership{Senior Member,~IEEE}
\thanks{
\textit{({Corresponding author: Rui Fan})} }
\thanks{The authors are with the College of Electronics \& Information Engineering, Shanghai Research Institute for Intelligent Autonomous Systems, the State Key Laboratory of Intelligent Autonomous Systems, and Frontiers Science Center for Intelligent Autonomous Systems, Tongji University, Shanghai 201804, China (e-mail: \{guosicen, tmi, cwliu, qjchen, rfan\}@tongji.edu.cn)}
}


\maketitle
\begin{abstract}
Recent vision foundation models (VFMs), typically based on Vision Transformer (ViT), have significantly advanced numerous computer vision tasks. Despite their success in tasks focused solely on RGB images, the potential of VFMs in RGB-depth driving scene parsing remains largely under-explored. In this article, we take one step toward this emerging research area by investigating a feasible technique to fully exploit VFMs for generalizable RGB-depth driving scene parsing. Specifically, we explore the inherent characteristics of RGB and depth data, thereby presenting a Heterogeneous Feature Integration Transformer (HFIT). This network enables the efficient extraction and integration of comprehensive heterogeneous features without re-training ViTs. Relative depth prediction results from VFMs, used as inputs to the HFIT side adapter, overcome the limitations of the dependence on depth maps. Our proposed HFIT demonstrates superior performance compared to all other traditional single-modal and data-fusion scene parsing networks, pre-trained VFMs, and ViT adapters on the Cityscapes and KITTI Semantics datasets. We believe this novel strategy paves the way for future innovations in VFM-based data-fusion techniques for driving scene parsing.
Our source code is publicly available at \url{https://mias.group/HFIT}.
\end{abstract}

\section{Introduction}
\label{sec:intro}

The rise of vision foundation models (VFMs) \cite{kirilloV2023segment,oquab2023dinoV2,yang2024depth,feng2023vipocc} has heralded a new era in computer vision. Traditional fundamental computer vision tasks, such as semantic segmentation \cite{kirilloV2023segment}, object detection \cite{han2024few} and depth estimation \cite{yang2024depth,liu2024playing}, have all started turning to VFMs for more compelling performance. This feasibility generally arises from the VFM's stronger capability in extracting profound, informative features through a range of related computer vision tasks \cite{kirilloV2023segment,oquab2023dinoV2,yang2024depth}. 
Despite several attempts made in RGB-depth (often abbreviated as RGB-D) driving scene parsing \cite{shvets2024joint,chen2024bridging}, existing prior arts based on VFMs are sketchy, and the achieved results remain unsatisfactory \cite{li2024hapnet}. Therefore, this article primarily focuses on fully exploiting VFMs' profound prior knowledge for RGB-D driving scene parsing.

\begin{figure}[t!]
\centering
\includegraphics[width=0.49\textwidth]{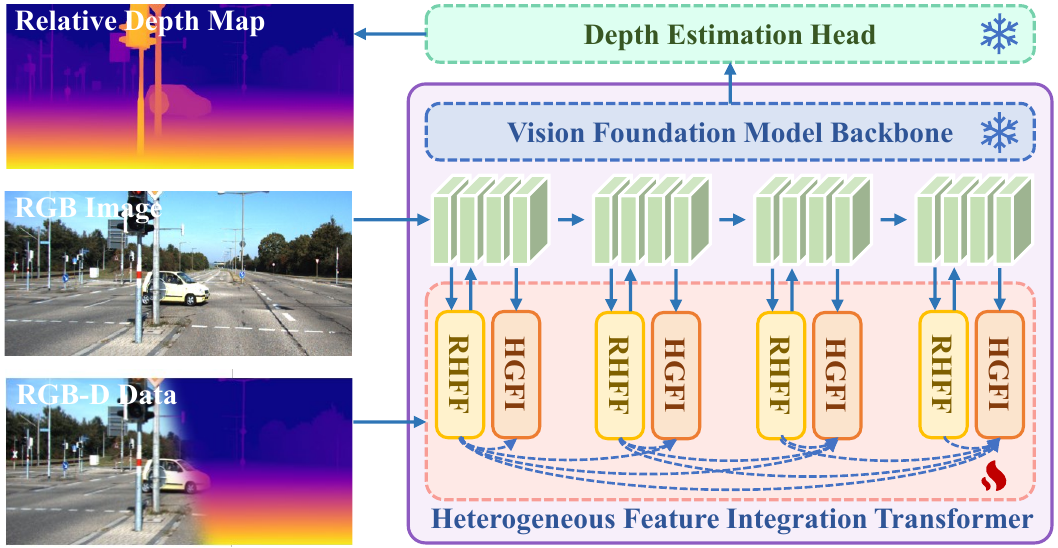}
\centering
\caption
{
An overview of our proposed {\textbf{HFIT}}. Through the interaction between recalibrated heterogeneous feature fusion (RHFF) modules and holistic gated feature integration (HGFI) modules, heterogeneous features are fully integrated with the profound prior features.
}
\label{fig.framework1}
\end{figure} 
The integration of RGB images and depth maps has been proven to significantly improve the performance of driving scene parsing \cite{fan2020sne-roadseg,min2022orfd,ha2017mfnet,hazirbas2017fusenet}. However, a significant limitation of these fusion models is their reliance on depth maps, which restricts their applicability in scenarios where LiDARs are unavailable. Stereo cameras provide a practical alternative for obtaining depth information \cite{fan2019pothole,fan2021graph}. However, even supervised methods fall short of achieving LiDAR-level accuracy, and the performance gap is even wider for unsupervised and self-supervised approaches \cite{liu2024these}.
Furthermore, integrating these heterogeneous characteristics can degrade scene parsing performance when the depth maps lack sufficient precision, \textit{e.g.}, due to inaccuracies in camera-LiDAR calibration \cite{wu2024s, huang2024online}.
Surprisingly, VFMs like Depth Anything V2 \cite{yang2024depth2} have demonstrated exceptional effectiveness not only in extracting informative deep features but also in learning geometric information and performing relative depth estimation. 
Since these methods perceive only relative spatial relationships and cannot directly derive exact depth values, many researchers assume that their impact on scene parsing is minimal \cite{hoyer2021three,saha2021learning}. However, we contend that relative depth maps are sufficient to enhance the performance of scene parsing. Utilizing VFMs' relative depth estimation output to assist data-fusion models deserves greater attention. Thus, this article marks the initial exploration in this uncharted territory, specifically focusing on the utilization of not only the VFMs' deep features but also their relative depth outputs. 

The main challenges for data-fusion models to have spatial understanding ability are in the following aspects.
First, VFMs have limited capacity to understand depth maps as they are trained only on RGB images \cite{oquab2023dinoV2,yang2024depth,yang2024depth2}. Consequently, directly inputting depth maps into VFMs results in poor performance. Most existing data-fusion models typically use dual encoders with separate trainable parameters to extract heterogeneous features from the RGB-D pairs \cite{fan2020sne-roadseg,min2022orfd,shvets2024joint}. Nevertheless, this strategy not only increases computational load but also overlooks the inherent characteristics between heterogeneous features. Therefore, developing an architecture that can leverage VFMs' informative prior knowledge while enabling the extraction and integration of spatial features with minimum resource utilization, stands as an under-explored area requiring further investigation. 
Additionally, RGB images are abundant in global semantic cues \cite{seichter2021efficient}, making them suitable for feature extraction via pre-trained Vision Transformers (ViTs), while depth images provide gradient-related semantics (local details) \cite{wang2016learning} which can be captured more efficiently through convolutional neural networks (CNNs). 
Thus, side adapters may be the optimal architecture choice. Freezing pre-trained ViTs and solely updating adapters not only leverages the profound prior knowledge of ViTs but also enables the model to extract spatial features with minimal resource usage \cite{chen2022vision,chen2024conv}. Although it may not achieve the performance of full fine-tuning strategies, its significantly lower number of trainable parameters and reduced resource consumption provide compelling advantages over full fine-tuning.

Additionally, strategies for fusing prior features from ViTs with heterogeneous features from the adapter are crucial for generalizable RGB-D driving scene parsing \cite{zhou2022canet,zhou2022fanet}. Indiscriminate fusion strategies, such as direct overlaying interactions between modalities \cite{ha2017mfnet}, not only introduce noise into the feature space but also exhibit the inadaptability of RGB-based models to mixed-modality inputs \cite{li2024roadformer,huang2024roadformer+}. Since traditional adapter inputs are typically RGB images, indiscriminate feature fusion \cite{chen2022vision, xia2024vit} is relatively harmless. However, if the adapter extracts heterogeneous information, indiscriminate fusion strategies will potentially drown out valid complementary information due to different semantic spaces \cite{wu2022robust,sun2021deep}. Therefore, it becomes essential to explore the inherent differences in heterogeneous feature characteristics, emphasizing their respective strengths and complementing each other based on reliability.

In addition to integrating heterogeneous features, the fusion of multi-level features with varying degrees of semantics has often been overlooked \cite{ha2017mfnet,seichter2021efficient,zhou2022canet,min2022orfd}. As the network depth increases, semantic clues become richer, while fine-grained information, such as boundary details, tends to be lost \cite{li2020gated}. SNE-RoadSeg \cite{fan2020sne-roadseg} and UNet++ \cite{zhou2019unet++} employ densely connected skip connections to enhance comprehensive holistic feature integration. While this approach incorporates both fine-grained and coarse-grained details, it fails to address the semantic gap between features extracted at different network depths. These limitations underscore the importance of assessing the significance of each feature vector within the feature map and aggregating the information accordingly.

To tackle the limitations discussed above, we introduce \uline{\textbf{H}eterogeneous \textbf{F}eature \textbf{I}ntegration \textbf{T}ransformer} (\textbf{HFIT}) (see Fig. \ref{fig.framework1}). HFIT investigates the utilization of VFMs for generalizable RGB-D driving scene parsing, thereby unleashing the power of profound prior knowledge embedded within these pre-trained VFMs. We begin by performing relative depth estimation \cite{yang2024depth2}. Then, we develop a side adapter to capture robust and multi-scale spatial pyramid features from RGB-D pairs and adaptively integrate them with the prior features from VFMs. HFIT adapter starts with a duplex spatial prior extractor (DSPE), which aims to capture local spatial priors from the RGB-D data. Then, we develop a recalibrated heterogeneous feature fusion module, which supplements VFMs with multi-scale spatial geometric features. Finally, we design a holistic gated feature integration module, which first selectively integrates features from multiple levels and then fuses multi-scale features from VFMs and the adapter.
Extensive experiments have been conducted on publicly available datasets \cite{cordts2016cityscapes,menze2015kitti}, and unequivocally demonstrate the superior performance of our proposed HFIT.
Our contributions are as follows:
\begin{itemize}
    \item To the best of our knowledge, this study is the first to investigate the adaptation of VFMs for RGB-D driving scene parsing. HFIT can directly extract and integrate heterogeneous features without re-training ViTs. By leveraging relative depth prediction results from VFMs as inputs to the HFIT adapter, our approach overcomes the limitations of the dependence on depth maps. 
    \item We introduce a DSPE strategy to jointly capture local semantics from RGB-D data, enabling HFIT to effectively learn heterogeneous features.
    \item We design an RHFF module to collaboratively recalibrate spatial priors, enhancing the comprehensiveness of feature fusion.
    \item We propose an HGFI strategy to combine complementary strengths of multi-level features, equipping HFIT with enhanced fine-grained feature representation capabilities.
\end{itemize}

\section{Related Work}
\label{sec:related_work}

\subsection{Conventional RGB-D Semantic Segmentation Networks}
\label{sec:RGB-D}
Considering the feature fusion stage of state-of-the-art (SoTA) RGB-D semantic segmentation networks, we can categorize them into three main types: early fusion, intermediate fusion, and late fusion \cite{zhang2021deep}. Early fusion methods typically combine RGB and depth images at the input stage. While this approach is straightforward, it falls short in achieving a comprehensive understanding of the environment \cite{zhang2021deep}. Intermediate fusion techniques \cite{fan2020sne-roadseg} extract and fuse these heterogeneous features from RGB-D pairs using duplex encoders. Late fusion approaches \cite{ha2017mfnet,valada2017adapnet,cheng2017locality} utilize two parallel encoders, with a primary focus on feature fusion within the decoder. Although these strategies have proven to be more effective than single-modal methods, they still face challenges such as insufficient utilization of depth features and high computational costs associated with multiple feature extractors \cite{huang2024roadformer+}. 

\subsection{Feature Extraction and Fusion Strategies}
\label{sec:feature}
Feature extraction and fusion play essential roles in advancing various computer vision tasks \cite{hassan2020learning,zhou2021ecffnet,jiang2013salient,zhang2018exfuse}. Traditional fusion strategies perform effectively when features exhibit relevance and consistency. However, when dealing with multi-scale, multi-level, and multi-modality features, indiscriminate extraction and fusion can introduce redundancies and noise due to the semantic gap \cite{wu2024s}. To address these limitations, the separation-and-aggregation gate (SAGate) \cite{chen2020bi} recalibrates and aggregates heterogeneous features to generate enhanced selective representations specifically for segmentation. Additionally, several studies \cite{fan2020sne-roadseg,zhou2019unet++} have addressed this issue by introducing densely connected skip connections to integrate fine-grained and coarse-grained information. The cross feature module (CFM) \cite{wei2020f3net} incorporates adaptive aggregation mechanisms to selectively retain useful feature maps and refine multi-level features while filtering out background noise. Nevertheless, these models often overlook the semantic gap between hierarchical features extracted at different network depths, which may result in incomplete fusion of high-level and low-level details. Hence, our primary focus in this article is on addressing this critical limitation.

\subsection{Vision Foundation Models}
\label{sec:DINOV2}
Vision foundation models have emerged as a groundbreaking approach in artificial intelligence, demonstrating superior learning capabilities compared to traditional models. The segment anything model (SAM) \cite{kirilloV2023segment} is a recently proposed VFM for image segmentation. Trained on a massive dataset consisting of over 1 billion masks from 11 million images, SAM is built to generalize across diverse segmentation tasks. By performing discriminative self-supervised learning at image and patch levels, DINOv2 \cite{oquab2023dinoV2} learns all-purpose visual features for various downstream tasks. Depth Anything V1 \cite{yang2024depth} inherits profound semantic priors from the pre-trained DINOv2 backbone via a simple feature alignment constraint, achieving superior performance in downstream tasks. Depth Anything V2 \cite{yang2024depth2} produces finer and more robust depth predictions than the first version by mitigating the distributional shift and limited diversity of synthetic data. Larger models naturally excel at integrating vision features, making them well-suited for our RGB-D driving scene parsing tasks. However, they also come with increased memory requirements and higher training costs. To strike a balance between performance, resource efficiency, and experimental versatility, we select Large DINOv2, Large Depth Anything V1, and Large Depth Anything V2 as the backbones for our proposed HFIT.

\section{Methodology}
\label{sec:methodology}

\subsection{Overall Workflow}
\label{sec:overview}

\begin{figure*}[t!]
\centering
\includegraphics[width=0.999\textwidth]{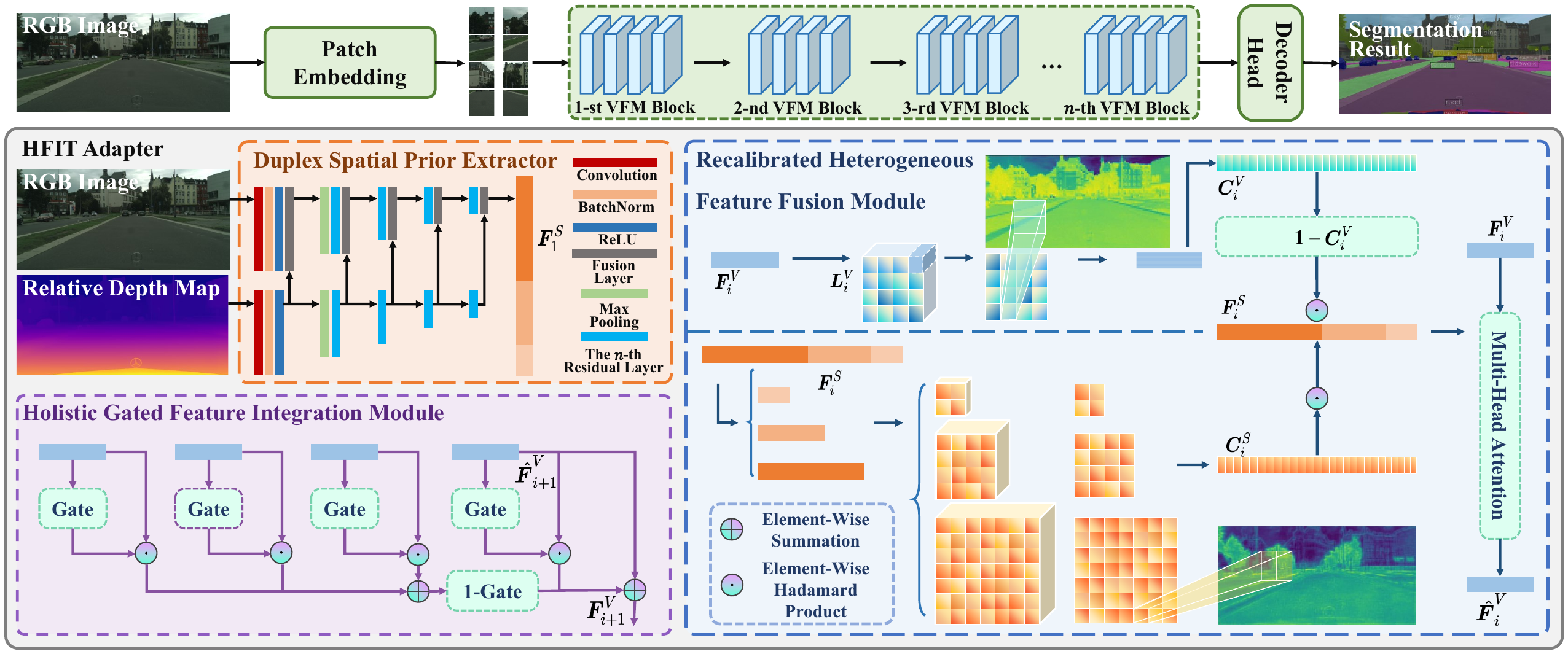}
\centering
\caption
{
An illustration of our proposed {\textbf{HFIT}}, consisting of (1) a plain ViT, (2) a duplex spatial prior extractor, (3) recalibrated heterogeneous feature fusion modules, and (4) holistic gated feature integration modules.
}
\label{fig.framework}
\end{figure*} 

As shown in Fig. \ref{fig.framework}, our HFIT comprises two parts: the first is a plain ViT, which consists of a patch embedding layer followed by $L$ encoder layers, and the second is our proposed HFIT adapter, which contains (1) a DSPE to capture local spatial features from the input RGB-D pairs, (2) RHFF modules to inject recalibrated spatial priors into the ViT, and (3) HGFI modules to integrate holistic hierarchical features and to reconstruct multi-level and multi-scale features.

First, in the ViT branch, the RGB image $\boldsymbol{I}^R \in \mathbb{R}^{ B \times 3 \times H\times W}$ is fed into the patch embedding layer to extract features with a resolution reduced to 1/16 of the original image. These patches are then flattened and projected into $D$-dimensional tokens, which, after being added with the position embedding, pass through $L$ encoder layers. Simultaneously, in the adapter branch, the RGB image and its corresponding depth map $\boldsymbol{I}^D \in \mathbb{R}^{ B \times 3 \times H\times W}$ pass through the DSPE to generate a feature pyramid $\left\{\boldsymbol{F}_{1,\frac{1}{8}}^S, \boldsymbol{F}_{1,\frac{1}{16}}^S, \boldsymbol{F}_{1,\frac{1}{32}}^S\right\}$. These pyramid features are then flattened and concatenated into $\boldsymbol{F}_1^S \in \mathbb{R}^{\left (\sum_{t=2}^4 \frac{H W}{S_t{ }^2}\right) \times D}$ as input for subsequent feature interaction, where $S_t=2^{t+1} (t \in[2,4] \cap \mathbb{Z})$ denotes the corresponding stride number. Specifically, given the number of interactions $N$ (usually $N$ = 4), we evenly split the Transformer encoders into $N$ blocks, each containing $L/N$ encoder layers. At the $i$-th stage, the spatial priors $\boldsymbol{F}_i^S$ are first injected into the Transformer features $\boldsymbol{F}_i^V \in \mathbb{R}^{\frac{H W}{16^2} \times D}$ via the RHFF module, which adaptively emphasizes and diminishes complementary information to achieve comprehensive fusion. Then, the HGFI module combines multi-level features and reorganizes multi-scale features from the ViT’s single-scale features at the end of each stage. After $N$ bidirectional feature interactions, the heterogeneous features from two branches are aggregated and passed to the back-end decoders for dense prediction tasks. 

\subsection{Duplex Spatial Prior Extractor}
\label{sec:DSPE}
Recent studies have shown that convolutions can help Transformers better capture local spatial semantics, underscoring their significance in computer vision tasks that necessitate clear object boundaries \cite{yuan2021incorporating}. Building on this idea, we propose the DSPE module, which is designed to capture local spatial contexts from RGB-D pairs in conjunction with the patch embedding layer. As depicted in Fig. \ref{fig.framework}, the DSPE employs dual branches to extract spatial priors from RGB images and relative depth maps. By incorporating RGB images as complementary inputs, the DSPE extracts more comprehensive local spatial patterns than using relative depth maps alone. EfficientNet \cite{tan2019efficientnet}, known for its superior performance in extracting detailed visual features, outperforms other hierarchical CNN models such as the ResNet \cite{he2016deep} series. Hence, we construct the DSPE using two identical EfficientNet blocks, enabling the efficient extraction of spatial priors from RGB-D pairs.

$\boldsymbol{I}^R \in \mathbb{R}^{ B \times 3 \times H\times W}$ and $\boldsymbol{I}^D \in \mathbb{R}^{ B \times 3 \times H\times W}$ are independently fed into duplex EfficientNet \cite{tan2019efficientnet} blocks, generating multi-scale feature pyramids $\mathcal{F}_1^R=\left\{\boldsymbol{F}_{1,\frac{1}{8}}^R, \boldsymbol{F}_{1,\frac{1}{16}}^R, \boldsymbol{F}_{1,\frac{1}{32}}^R\right\}$ and 
$\mathcal{F}_1^D=\left\{\boldsymbol{F}_{1,\frac{1}{8}}^D, \boldsymbol{F}_{1,\frac{1}{16}}^D, \boldsymbol{F}_{1,\frac{1}{32}}^D\right\}$, where $\boldsymbol{F}_{1,\frac{1}{2^i}}^{R, D} \in \mathbb{R}^{\frac{H}{S_l} \times \frac{W}{S_l} \times C_l}$ represents the features at the $l$-th stage, $C_l$ and $S_l=2^{l+1}(l \in[2,4] \cap \mathbb{Z})$ denote the corresponding channel and stride numbers, respectively. 
Subsequently, $\boldsymbol{F}_{1,\frac{1}{2^l}}^R$ and $\boldsymbol{F}_{1,\frac{1}{2^l}}^D$ are merged via element-wise summation, followed by $1 \times 1$ convolution layers to reduce the output to $D$ channels, yielding heterogeneous features $\mathcal{F}_1^S=\left\{\boldsymbol{F}_{1,\frac{1}{8}}^S, \boldsymbol{F}_{1,\frac{1}{16}}^S, \boldsymbol{F}_{1,\frac{1}{32}}^S\right\}$, where $\boldsymbol{F}_{1,\frac{1}{2^t}}^S \in \mathbb{R}^{\frac{H}{S_t} \times \frac{W}{S_t} \times D}$. 
These three-scale features are then flattened and concatenated to form the heterogeneous spatial prior $\boldsymbol{F}_1^S \in \mathbb{R}^{\left(\sum_{t=2}^4 \frac{H W}{S_t{ }^2}\right) \times D}$, which is utilized in subsequent heterogeneous feature fusion stages.

\subsection{Recalibrated Heterogeneous Feature Fusion Module}
\label{sec:awrfm}
The primary challenge in feature extraction for RGB-D scene parsing lies in the efficient fusion and integration of diverse heterogeneous features obtained from multiple data sources \cite{wang2016learning,sun2020real,xu2023thcanet}. As discussed in Sect. \ref{sec:related_work}, previous studies \cite{ha2017mfnet,zhou2022canet,zhou2022fanet} combined these heterogeneous features without adequately addressing their inherent differences. To tackle these limitations, RHFF selectively highlights key regions while diminishing the focus on less significant areas. Original heterogeneous spatial priors are adaptively recalibrated through dynamic weighting and seamlessly integrated into the ViT via a multi-scale deformable attention mechanism. 

Before injecting heterogeneous features, it is crucial to distinguish and quantify the reliability of each feature. Convolutional layers are initially employed to generate a logit map $\boldsymbol{L}_i^V \in \mathbb{R}^{\frac{H}{16} \times \frac{W}{16} \times C}$ for Transformer features through the following process: 
\begin{equation}
\boldsymbol{L}_i^V=\operatorname{ReLU}\left(\operatorname{BN}\left(\underset{1 \times 1}{\operatorname{Conv}}\left(\operatorname{Reshape} (\boldsymbol{F}_i^V)\right)\right)\right),
\end{equation}
where ``$\operatorname{BN}(\cdot)$'' denotes BatchNorm operation, ``$\operatorname{Reshape}$'' operation reshapes the two-dimensional matrix $\boldsymbol{F}_i^V\in \mathbb{R}^{\frac{HW}{16^2} \times D}$ into a three-dimensional tensor $\boldsymbol{F}_i^V\in \mathbb{R}^{\frac{H}{16} \times \frac{W}{16} \times D}$, and $C$ denotes the segmentation classes. After modeling interdependencies across channel dimensions, the reliabilities are contrasted to compute the confidence map $\boldsymbol{C}_{i,\frac{1}{16}}^V \in \mathbb{R}^{\frac{H}{16} \times \frac{W}{16} \times 1}$ using the following expression:
\begin{equation}
\boldsymbol{C}_{i,\frac{1}{16}}^V=\sigma\left({\underset{k\times k,r}{\operatorname{Conv}}}\left(-\boldsymbol{L}_i^V \odot \log \boldsymbol{L}_i^V\right)\right),
\end{equation}
where $\odot$ represents the element-wise Hadamard  product between two metrics, $\sigma(\cdot)$ denotes the Sigmoid operation, and  ``$\underset{k\times k,r}{\operatorname{Conv}}$'' denotes atrous convolutional layers with kernel size $k$ and dilation rate $r$. 
Each element in $\boldsymbol{C}_i^V$ indicates the confidence level of a pixel relative to its segmentation class. 
A higher confidence level suggests the pixel significantly contributes to accurate segmentation, while a lower confidence level implies redundancy, which should be either ignored or supplemented with relevant information.
To align the feature pyramid $\boldsymbol{F}_i^S \in \mathbb{R}^{\left(\sum_{t=2}^4 \frac{H W}{S_t{ }^2}\right) \times D}$ across three resolutions, we downsample and upsample $\boldsymbol{C}_{i,\frac{1}{16}}^V$ to $\boldsymbol{C}_{i,\frac{1}{8}}^V \in \mathbb{R}^{\frac{H}{8} \times \frac{W}{8} \times 1}$ as well as $\boldsymbol{C}_{i,\frac{1}{32}}^V \in \mathbb{R}^{\frac{H}{32} \times \frac{W}{32} \times 1}$. These features are then concatenated to yield the confidence map $\boldsymbol{C}_i^V$ as follows:
\begin{equation}
\boldsymbol{C}_i^V=\left[\boldsymbol{C}_{i,\frac{1}{8}}^V ; \boldsymbol{C}_{i,\frac{1}{16}}^V; \boldsymbol{C}_{i,\frac{1}{32}}^V\right] \in \mathbb{R}^{\left(\sum_{t=2}^4 \frac{H W}{S_t{ }^2}\right) \times 1}.
\end{equation}

In the adapter branch, the spatial prior $\boldsymbol{F}_i^S \in \mathbb{R}^{\left(\sum_{t=2}^4 \frac{H W}{S_t{ }^2}\right) \times D}$ is first divided into three-scale feature maps $\boldsymbol{F}_{i,\frac{1}{2^t}}^S \in \mathbb{R}^{\frac{H}{S_t} \times \frac{W}{S_t} \times D}$. Then, we calculate the confidence map of the heterogeneous spatial prior $\boldsymbol{C}_i^S$ as follows:
\begin{equation}
\boldsymbol{L}_{i,\frac{1}{2^t}}^S=\operatorname{ReLU}\left(\operatorname{BN}\left(\underset{1 \times 1}{\operatorname{Conv}}\left(\operatorname{Reshape} (\boldsymbol{F}_{i,\frac{1}{2^t}}^S)\right)\right)\right),
\end{equation}
\begin{equation}
\boldsymbol{C}_{i,\frac{1}{2^t}}^S=\operatorname{Reshape}'\left(\sigma \left({\underset{k\times k,r}{\operatorname{Conv}}}\left(-
\boldsymbol{L}_{i,\frac{1}{2^t}}^S \odot \log \boldsymbol{L}_{i,\frac{1}{2^t}}^S\right)\right)\right),
\end{equation}
\begin{equation}
\boldsymbol{C}_i^S=\left[
\boldsymbol{C}_{i,\frac{1}{8}}^S ; 
\boldsymbol{C}_{i,\frac{1}{16}}^S; 
\boldsymbol{C}_{i,\frac{1}{32}}^S
\right] \in \mathbb{R}^{\left(\sum_{t=2}^4 \frac{H W}{S_t{ }^2}\right) \times 1},
\end{equation}
where “$\operatorname{Reshape}'$” expands the three-dimensional tensor into a two-dimensional matrix.
In contrast to previous works like SAGate \cite{chen2020bi}, CFM \cite{wei2020f3net}, and DDPM \cite{pang2020hierarchical}, which mainly focus on feature commonality, our RHFF module delves deeper into complementary and redundant features.
Since a plain ViT may struggle to confidently segment certain regions, these areas should be supplemented with heterogeneous spatial priors. Additionally, these priors need to be validated for reliability to avoid introducing redundant or incorrect information.
Upon such inspirations, we use these two confidence maps to adaptively recalibrate the heterogeneous spatial priors as follows:
\begin{equation}
\boldsymbol{F}_i^S=\left(1-\boldsymbol{C}_i^V\right) \odot \boldsymbol{C}_i^S \odot \boldsymbol{F}_i^S,
\end{equation}
where $(1-\boldsymbol{C}_i^V)$ selectively activates spatial priors that are deficient in Transformer features, and $\boldsymbol{C}_i^S$ emphasizes high-reliability regions within spatial priors.

After adaptively recalibrating and purifying the heterogeneous spatial priors, we utilize a cross-attention mechanism to inject $\boldsymbol{F}_i^S$ into the Transformer feature $\boldsymbol{F}_i^V$. The process can be described as:
\begin{equation}
\hat{\boldsymbol{F}}_i^V=\boldsymbol{F}_i^V+\boldsymbol{\gamma}_i \operatorname{MHA}\left(\operatorname{LN}\left(\boldsymbol{F}_i^V\right), \operatorname{LN}\left(\boldsymbol{F}_i^S\right)\right),
\end{equation}
where ``$\operatorname{LN}(\cdot)$'' denotes LayerNorm operation \cite{lei2016layer}, and ``$\operatorname{MHA}(\cdot)$'' is a multi-head attention operation \cite{cheng2022masked}. In addition, we introduce a learnable vector $\boldsymbol{\gamma}^i \in \mathbb{R}^D$ \cite{huang2021fapn,xia2024vit} to ensure that the feature distribution of $\boldsymbol{F}_i^V$ remains largely unchanged. This allows for effective utilization of the pre-trained weights of the ViT while maintaining stable feature integration.

\subsection{Holistic Gated Feature Integration Module}
\label{sec:HGFI}
Integrating low-level features to recover detailed information lost in high-level features is a natural approach \cite{fu2013integrating,zhang2018exfuse}. Unfortunately, directly combining them suffers from the semantic gap and will drown useful information \cite{li2020gated}. To this end, we propose a holistic integration module to selectively extract features from multiple levels. Specifically, lower-level features, rich in fine-grained details, are used to enhance features at higher levels, while gate mechanisms regulate the flow of relevant information.

After injecting heterogeneous spatial priors into the plain ViT, we obtain the output feature $\hat{\boldsymbol{F}}_{i+1}^V$ by passing $\hat{\boldsymbol{F}}_i^V$ through the backbone layers of the $i$-th block. Then, we utilize the gate mechanism to control the feature flow as follows:
\begin{equation}
\begin{aligned}
\boldsymbol{F}_{i+1}^V=\big(1+\boldsymbol{G}_{i+1}^V\big) \odot \hat{\boldsymbol{F}}_{i+1}^V
+\big(1-\boldsymbol{G}_{i+1}^V\big) \odot \sum_{l=1}^i \boldsymbol{G}_{l}^V \odot \hat{\boldsymbol{F}}_{l}^V,
\end{aligned}
\end{equation}
\begin{equation}
\begin{aligned}
\boldsymbol{F}_{i}^S=\big(1+\boldsymbol{G}_{i}^S\big) \odot {\boldsymbol{F}}_{i}^S
+\big(1-\boldsymbol{G}_{i}^S\big) \odot \sum_{l=1, l \neq i}^i \boldsymbol{G}_{l}^S \odot {\boldsymbol{F}}_{l}^S.
\end{aligned}
\end{equation}
Features from lower levels are propagated to the $i$-th stage only when $\boldsymbol{G}_{l}^{V,S}$ has high values and $\boldsymbol{G}_{i}^{V,S}$ has low values. In addition to regulating the flow of useful information to appropriate stages, the gates can effectively prevent information redundancy by updating features only when the current information is insufficient or irrelevant. $\boldsymbol{G}_{l}^V \in \mathbb{R}^{\frac{HW}{16^2} \times D}$ and $\boldsymbol{G}_{l}^S \in \mathbb{R}^{\left(\sum_{t=2}^4 \frac{H W}{S_t{ }^2}\right) \times D}$ are calculated as follows:
\begin{equation}
\boldsymbol{G}_{l}^{V} = \operatorname{Reshape}'\left(\sigma\left(\underset{1 \times 1}{\operatorname{Conv}}\left(\operatorname{Reshape}\left(\hat{\boldsymbol{F}}_{l}^V\right)\right)\right)\right),
\end{equation}
\begin{equation}
\boldsymbol{G}_{l,\frac{1}{2^t}}^{S} = \operatorname{Reshape}'\left(\sigma\left(\operatorname{DW}\left(\operatorname{Reshape}\left(\boldsymbol{F}_{i,\frac{1}{2^t}}^S\right)\right)\right)\right),
\end{equation}
\begin{equation}
\boldsymbol{G}_i^S=\left[
\boldsymbol{G}_{i,\frac{1}{8}}^S ; 
\boldsymbol{G}_{i,\frac{1}{16}}^S; 
\boldsymbol{G}_{i,\frac{1}{32}}^S
\right],
\end{equation}
where ``$\operatorname{DW}(\cdot)$'' represents a set of depth-wise convolutions with varying kernel sizes, designed to accommodate the diverse receptive field requirements of different feature groups.

Moreover, semantic representation often exhibits biases due to modality differences. To alleviate this problem, we employ a module comprising a cross-attention layer and a feed-forward network (FFN) to unify spatial priors with Transformer features. This process can be formulated as:
\begin{equation}
\boldsymbol{F}_{i+1}^S=\hat{\boldsymbol{F}}_{i}^S+\operatorname{FFN}\left(\operatorname{LN}\left(\hat{\boldsymbol{F}}_{i}^S\right)\right), 
\end{equation}
\begin{equation}
\hat{\boldsymbol{F}}_{i}^S=\boldsymbol{F}_{i}^S+\operatorname{MHA}\left(\operatorname{LN}\left(\boldsymbol{F}_{i}^S\right), \operatorname{LN}\left(\boldsymbol{F}_{i+1}^V\right)\right).
\end{equation}
Moreover, unlike traditional Transformer architecture \cite{liu2021swin,xie2021segformer}, which employs the attention mechanism solely on single-scale features, our HGFI strategy incorporates multi-resolution features at scales of 1/8, 1/16, and 1/32. The generated feature $\boldsymbol{F}_{i+1}^S$ is then passed to the next RHFF module.

\section{Experiments}
\label{sec:experiments}
\subsection{Datasets, Implementation Details, and Evaluation Metrics}
\label{sec.implement}
We utilize two public datasets to evaluate the performance of our proposed methods: Cityscapes \cite{cordts2016cityscapes} and KITTI Semantics \cite{menze2015kitti} datasets. Our model was trained for 20,000 iterations on an NVIDIA RTX 3090 GPU.
Images were randomly cropped to a size of 448$\times$448 pixels during the training process. Standard data augmentation techniques, such as random color adjustments, photometric distortion, rescaling, and flipping, were also used. We employed five widely used metrics to evaluate the scene parsing performance: mean F$_1$-score (mFsc), mean intersection over union (mIoU), average accuracy (aAcc), mean precision (mPre), and mean recall (mRec). More details are given in the supplementary material.

\begin{table*}[!t]
\begin{center}
\settablefont
\caption{Quantitative comparisons with SoTA scene parsing methods on the Cityscapes dataset.} 
\label{tb.sota_c}
{
\begin{tabular}{c|c|l|cccccc}
\toprule
Backbone
&Type
&Methods
&mFsc (\%) $\uparrow$  &mIoU (\%) $\uparrow$ &aAcc (\%) $\uparrow$
&mPre (\%) $\uparrow$  &mRec (\%) $\uparrow$ &error range (\%) $\downarrow$\\
\hline
\hline
\multirow{7}{*}{\rotatebox[origin=c]{90}{Traditional}} 
&\multirow{3}{*}{Single-Modal} 
&OCRNet \cite{yuan2020object} &79.78   &67.29   &94.99    &79.73   &77.91 & 1.23\\
&&KNet \cite{zhang2021k} &74.43   &62.25   &93.50    &81.58   &71.74 & 0.59\\
&&EMANet \cite{li2019expectation} &72.10   &61.52   &93.91    &78.85   &69.69 & 0.67\\
\cline{2-9}
&\multirow{4}{*}{Data-Fusion} 
&SNE-RoadSeg \cite{fan2020sne-roadseg} &54.04  &42.52  &89.34   &67.01  &49.78 & 2.32\\
&& OFFNet \cite{min2022orfd}&59.22  &49.16  &64.33   &71.21  &53.41& 1.54\\
&&MFNet \cite{ha2017mfnet}&51.50  &39.20  &89.39  &62.69  &46.56 & 0.81\\
&&FuseNet \cite{hazirbas2017fusenet} &53.28  &40.40   &89.79  &63.63 &49.57 & 2.03\\
\cline{1-9}
\multirow{8}{*}{\rotatebox[origin=c]{90}{VFM}} 
&\multirow{3}{*}{\makecell[c]{Single-Modal \\w/o Adapter}} 
& DINOv2 \cite{oquab2023dinoV2} &88.11  &80.23  &96.12  &89.29   &88.91& 0.52\\
&&Depth Anything V1 \cite{yang2024depth} &89.48  &81.11  &96.30  &90.26 &88.66 & \textbf{0.29}\\
&&Depth Anything V2 \cite{yang2024depth2} &89.00  &80.31   &96.03  &89.51  &88.27& 0.41\\
\cline{2-9}
&\multirow{2}{*}{\makecell[c]{\makecell[c]{Single-Modal \\w/ Adapter}}} 
& ViT-Adapter \cite{chen2022vision} &89.69  &82.11  &96.53  &90.91   &89.31& 0.35\\
&
&ViT-CoMer \cite{xia2024vit} &88.84  &80.02  &96.15  &89.21 &88.01 & 0.61\\
\cline{2-9}
&\multirow{3}{*}{\makecell[c]{Data-Fusion \\w/ Adapter (\textbf{ours})}} 
& DINOv2 \cite{oquab2023dinoV2} & 91.32  & 84.53  & 97.15  &  91.42   & \textbf{91.19}& 0.42\\
&
& Depth Anything V1 \cite{yang2024depth} & \textbf{91.64}  & \textbf{84.74}  & \textbf{97.09}  & \textbf{92.27} & 90.26 & 0.64\\
& 
& Depth Anything V2 \cite{yang2024depth2} & 91.06  & 84.59   & 96.94  & 91.87  & 90.49& 0.52\\
\bottomrule
\end{tabular}}
\end{center}
\end{table*}
\begin{table*}[!t]
\begin{center}
\settablefont
\caption{Quantitative comparisons with SoTA scene parsing methods on the KITTI Semantics dataset.} 
\label{tb.sota_s}
{
\begin{tabular}{c|c|l|cccccc}
\toprule
Backbone
&Type
&Methods
&mFsc (\%) $\uparrow$  &mIoU (\%) $\uparrow$ &aAcc (\%) $\uparrow$
&mPre (\%) $\uparrow$  &mRec (\%) $\uparrow$ &error range (\%) $\downarrow$\\
\hline
\hline
\multirow{7}{*}{\rotatebox[origin=c]{90}{Traditional}} 
&\multirow{3}{*}{Single-Modal} 
&OCRNet \cite{yuan2020object} &79.77   &62.24   &93.26    &81.40   &68.70 & 0.54\\
&&KNet \cite{zhang2021k} &74.56   &56.47   &91.22    &84.74   &62.79 & 0.72\\
&&EMANet \cite{li2019expectation} &67.39   &57.03   &90.42    &80.28   &63.71 & 0.67\\
\cline{2-9}
&\multirow{4}{*}{Data-Fusion} 
&SNE-RoadSeg \cite{fan2020sne-roadseg} &67.74  &41.28  &89.92   &70.39  &50.81& 1.03\\
&& OFFNet \cite{min2022orfd}&55.12  &39.24  &91.65   &67.47  &46.26& 1.23\\
&&MFNet \cite{ha2017mfnet}&58.55  &37.97  &88.65  &69.70  &42.98 & 0.64\\
&&FuseNet \cite{hazirbas2017fusenet}&50.91  &34.12   &84.38  &50.94 &42.68 & 0.75\\
\cline{1-9}
\multirow{8}{*}{\rotatebox[origin=c]{90}{VFM}} 
&\multirow{3}{*}{\makecell[c]{Single-Modal \\w/o Adapters}} 
& DINOv2 \cite{oquab2023dinoV2}&86.82  &78.63  &95.49 &87.84   &86.39& 0.34\\
&&Depth Anything V1 \cite{yang2024depth}  &86.62  &77.85  &95.63  &89.41 &84.66 & \textbf{0.22}\\
&&Depth Anything V2 \cite{yang2024depth2}  &87.05  &79.30   &95.77  &89.61  &86.04& 0.45\\
\cline{2-9}
&\multirow{2}{*}{\makecell[c]{\makecell[c]{Single-Modal \\w/ Adapters}}} 
& ViT-Adapter \cite{chen2022vision} &87.28 &79.28  &95.91  &88.27   &87.23& 0.23\\
&
&ViT-CoMer \cite{xia2024vit} &85.13  &76.54  &95.47  &88.93 &83.77 & 0.44\\
\cline{2-9}
&\multirow{3}{*}{\makecell[c]{Data-Fusion \\w/ Adapters (\textbf{ours})}} 
& DINOv2 \cite{oquab2023dinoV2}&87.58  &79.91 &\textbf{96.22}  &\textbf{89.41}  &86.88& 0.52\\
&
& Depth Anything V1 \cite{yang2024depth}  &\textbf{88.07}  &\textbf{80.24}  &95.82  &89.37 &\textbf{87.27} & 0.51\\
& 
& Depth Anything V2 \cite{yang2024depth2} &87.23  &79.38 &95.88  &89.13  &86.25& 0.63\\
\bottomrule
\end{tabular}}
\end{center}
\end{table*}

\subsection{Comparisons with State-of-the-Art Methods}
\label{sec.compare_with_sota}
\begin{figure*}[h]
\centering
\includegraphics[width=0.999\textwidth]{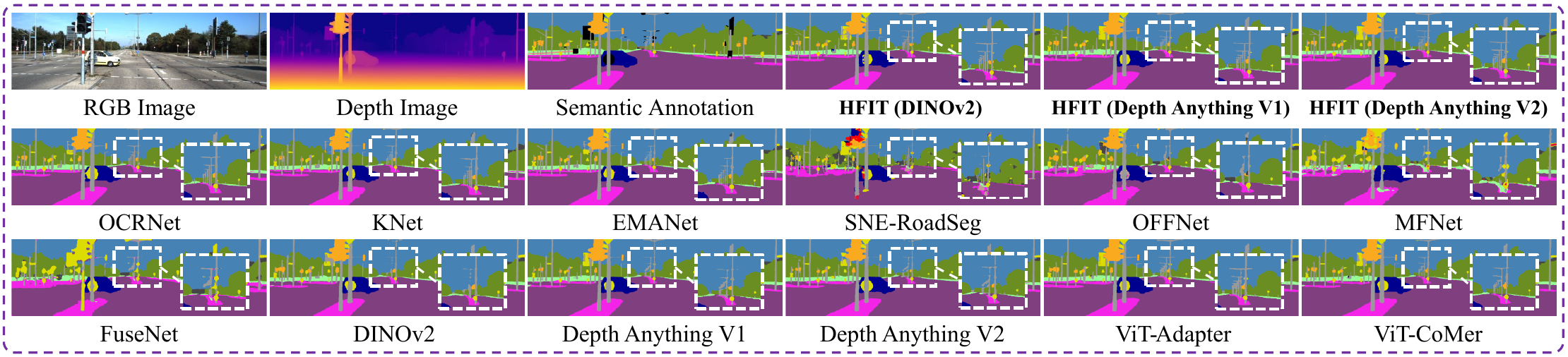}
\centering
\caption{Qualitative comparisons with SoTA scene parsing approaches on the KITTI Semantics \cite{menze2015kitti} dataset.}
\label{fig.kitti}
\end{figure*} 
\begin{figure*}[!t]
\centering
\includegraphics[width=0.999\textwidth]{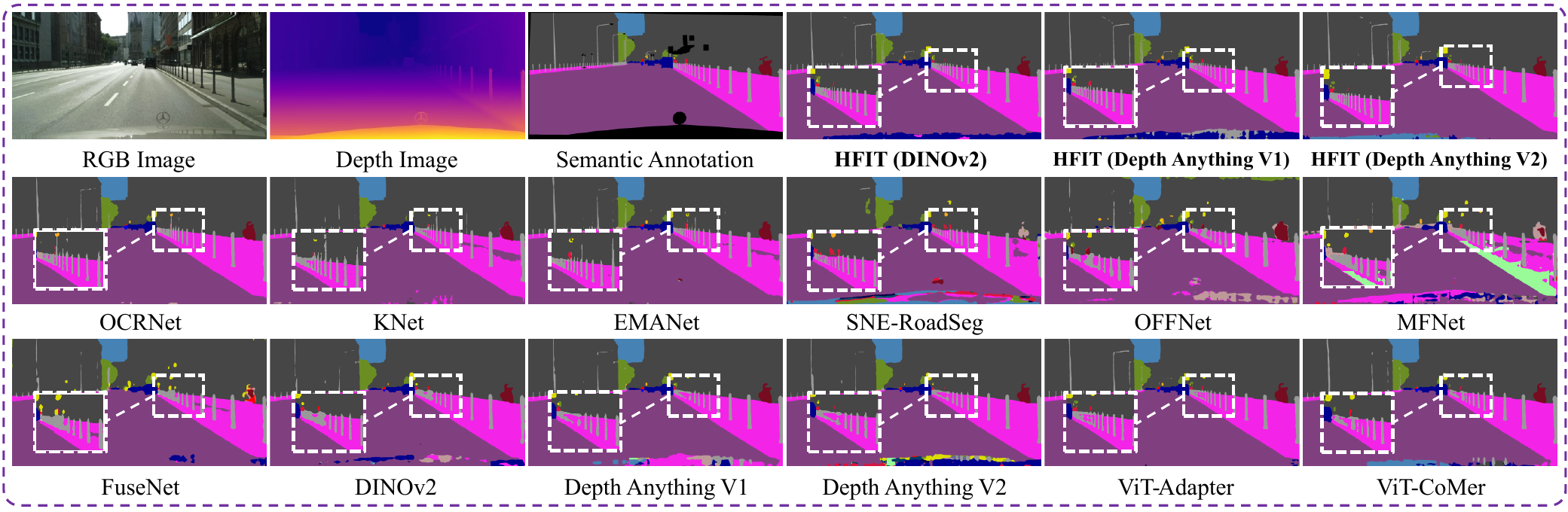}
\captionof{figure}{Qualitative comparisons with SoTA scene parsing approaches on the Cityscapes \cite{cordts2016cityscapes} dataset.}
\label{fig.cityscapes}
\end{figure*} 

As illustrated in Figs. \ref{fig.cityscapes} and \ref{fig.kitti}, HFIT consistently outperforms other scene parsing methods on the Cityscapes and KITTI Semantics datasets. As depicted in Figs. \ref{fig.cityscapes} and \ref{fig.kitti}, our HFIT achieves segmentation results with clear boundaries, effectively separating adjacent objects such as vehicles, pedestrians, and buildings. Additionally, HFIT demonstrates significant improvements in segmenting small and distant categories, including streetlights, traffic signs, and fences. In dense, visually complex regions like intersections, HFIT ensures robust semantic consistency, accurately segmenting objects such as sidewalks, vegetation, and poles, while minimizing overlap and misclassification.

Tables \ref{tb.sota_s} and \ref{tb.sota_c} highlight the best results in bold, with $\uparrow$ indicating that higher values correspond to improved performance. To ensure a thorough evaluation, we conduct additional experiments by averaging the results from three independent runs for each metric. These results indicate that our proposed HFIT outperforms all other traditional RGB-D methods and single-modal methods, with an increase in mIoU by 17.45-45.54\% on the Cityscapes dataset and an increase in mIoU by up to  46.12\% on the KITTI Semantics dataset. Compared to VFM-based algorithms, HFIT demonstrates greater performance, delivering a 2.63-4.72\% increase in mIoU on the Cityscapes dataset and a 0.94-2.39\% improvement on the KITTI dataset. These results underscore HFIT's superior visual representation capabilities and reliability in real-world driving scenarios. They also validate the hypothesis proposed in Sect. \ref{sec:intro}, demonstrating that incorporating heterogeneous spatial information significantly enhances the scene parsing performance of VFMs.

Compared to traditional RGB-D methods, single-modal VFMs demonstrate significant advantages, largely due to their profound prior knowledge, which enables them to achieve competitive results in constrained settings. Additionally, we select ViT-Adapter \cite{chen2022vision} and ViT-CoMer \cite{xia2024vit} as the baselines of single-modal VFMs with adapters. ViT-Adapter enhances ViT’s ability to capture multi-scale information by adding spatial prior, injector, and extractor modules. ViT-CoMer efficiently handles multi-scale interactions, balancing complex scene comprehension with memory efficiency.
Notably, ViT-Adapter \cite{chen2022vision} outperforms standard VFMs, attributed to its dynamic fine-tuning mechanism, which allows for better refinement of feature representations. Nevertheless, ViT-CoMer's \cite{xia2024vit} performance is underwhelming. This may be due to its bidirectional summation feature fusion mechanism, which fails to bridge the semantic gap between two distinct feature types. 

Additionally, we also validate the compatibility of HFIT with various cutting-edge VFMs. Among them, Depth Anything V1 \cite{yang2024depth} achieves the best performance, with the highest mIoU of 84.74\%. However, an interesting observation emerged during our experiments. Despite the evaluation matrices having comparable values on the Cityscapes \cite{cordts2016cityscapes} and KITTI Semantics \cite{menze2015kitti} datasets, the performance improvements achieved by HFIT compared to VFM-based methods are relatively less significant on the KITTI Semantics dataset. This performance gap can be attributed to KITTI’s smaller dataset size and limited diversity, which restrict the ability of large models like HFIT to fully utilize their representational capacity. As a result, the model is more prone to overfitting and exhibits reduced generalization on the KITTI dataset.

\begin{table}[!t]
\begin{center}
\settablefont
\caption{Ablation Study on DSPE with different inputs.
} 
\label{tb.DSPE}
{
\begin{tabular}{l|ccccc}
\toprule
Input Type
&mFsc (\%) $\uparrow$  &mIoU (\%) $\uparrow$ &aAcc (\%) $\uparrow$ \\
\hline
\hline
RGB   &90.39  &83.07   & 96.76   \\
Depth  &90.53 &83.85  &96.74   \\
Relative Depth   &90.42 &83.82  &96.78   \\
RGB+Depth  &91.02 &84.10  &\textbf{96.99}  \\
RGB+Relative Depth  &\textbf{91.11} &\textbf{84.22}  &96.97   \\

\bottomrule
\end{tabular}}
\end{center}
\end{table}

\begin{table}[!t]
\begin{center}
\settablefont
\caption{Ablation Study on DSPE blocks.} 
\label{tb.DSPE2}
{
\begin{tabular}{l|c|ccc}
\toprule
Input Type & Params (M)
&mFsc (\%) $\uparrow$  &mIoU (\%) $\uparrow$ &aAcc (\%) $\uparrow$ \\
\hline
\hline
ResNet-18  &23.34 &90.55  &83.31 &96.87  \\
ResNet-34  &43.56 &90.78  &83.68   &96.90 \\
ResNet-50  &50.95 &90.81  &83.76 &96.90    \\
\hline
Swin-T  &56.51 &90.76  &83.65   &96.86\\
Swin-S  &99.15 &90.84  &83.78   &96.88 \\
Swin-B  &175.45 &90.69  &83.56   &96.90 \\
\hline
EfficientNet-B0  &6.35 &90.53  &83.32  &96.81\\
EfficientNet-B1  &9.21 &90.69  &83.59 & \textbf{96.91}   \\
EfficientNet-B7  &79.09 &\textbf{90.85} &\textbf{83.82}    &96.89\\
\bottomrule
\end{tabular}
}
\end{center}
\end{table}

\subsection{Ablation Studies}
\label{sec.abla}

\begin{table}[!t]
\begin{center}
\settablefont
\caption{Ablation Study on RHFF Modules.
} 
\label{tb.fhff}
{
\begin{tabular}{c|c|ccc}
\toprule
RGB Weight & Depth Weight
&mFsc (\%) $\uparrow$  &mIoU (\%) $\uparrow$ &aAcc (\%) $\uparrow$ \\
\hline
\hline
&  &90.24  &82.83  &96.79   \\
$\checkmark$  & &90.31  &82.98  &96.77   \\
    &$\checkmark$ &90.43  &83.14    &96.83\\
$\checkmark$  & $\checkmark$ &\textbf{90.96}  &\textbf{84.01} & \textbf{96.97}   \\
\bottomrule
\end{tabular}
}
\end{center}
\end{table}

\begin{table}[t]
\begin{center}
\settablefont
\caption{Ablation Study on HGFI Modules.
} 
\label{tb.hgfi}
{
\begin{tabular}{c|c|ccc}
\toprule
HGFI-ViT & HGFI-Adapter
&mFsc (\%) $\uparrow$  &mIoU (\%) $\uparrow$ &aAcc (\%) $\uparrow$ \\
\hline
\hline
&  &89.91  &82.36  &96.73   \\
$\checkmark$  & &90.44  &83.17  &96.80   \\
    &$\checkmark$ &90.30  &82.96  &96.77   \\
$\checkmark$  & $\checkmark$ &\textbf{91.04}  &\textbf{84.11} &\textbf{96.94}   \\
\bottomrule
\end{tabular}
}
\end{center}
\end{table}

As shown in Table \ref{tb.DSPE}, we first evaluate HFIT’s performance using different input strategies. Single-modal configurations, using either RGB or depth images alone, achieve mIoU scores of 83.07\% and 83.85\%, respectively. Inputting both RGB and depth images achieves superior performance, with an mIoU of 84.10\%, validating the importance of modality complementation. Notably, the performance difference between using scaled metric depth obtained from RAFT-Stereo \cite{9665883} and relative depth derived from the VFM is minimal, within a margin of 0.12\%. This observation further supports the hypothesis presented in  Sect. \ref{sec:intro}.

\begin{figure*}[!t]
    \centering
    \includegraphics[width=0.999\textwidth]{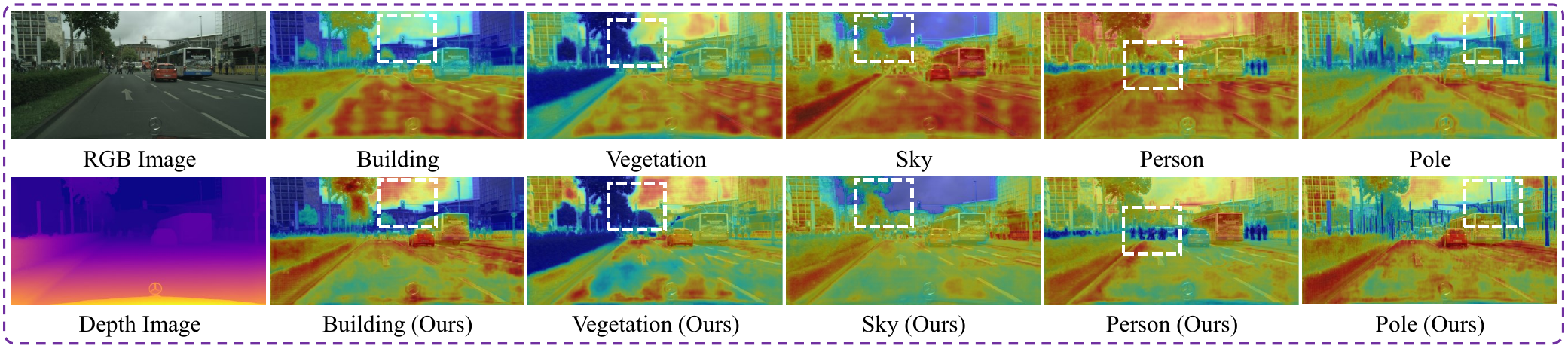}
    \captionof{figure}{Probability maps for different classes generated by VFMs and HFIT, where blue indicates high prediction confidence and red represents low prediction confidence.}
    \label{fig.feature}
\end{figure*}

To investigate the most effective structure for the DSPE, we substitute the basic blocks with alternatives, as detailed in Table \ref{tb.DSPE2}. 
Compared to earlier models like the ResNet \cite{he2016deep} series and the more recent Swin Transformer \cite{liu2021swin} series, EfficientNet \cite{tan2019efficientnet} demonstrates superior performance, achieving the highest mIoU of 83.82\%. This suggests that the lightweight design of EfficientNet is well-suited for effectively extracting spatial priors.

Theoretically, higher confidence and greater complementarity in features should result in improved fusion. To validate this hypothesis, we experimentally evaluate the contributions of the RGB weights $(1-\boldsymbol{C}_i^V)$ and depth weights $\boldsymbol{C}_i^S$ in the RHFF module. As expected, removing either component leads to a significant decline in HFIT's performance (see Table \ref{tb.fhff}). Moreover, we compare the probability maps produced by VFMs and HFIT. As shown in Fig. \ref{fig.feature}, the probability maps generated by HFIT demonstrate more reliable and confident predictions for each category, resulting in clearer and sharper probability distributions. This improvement is particularly evident for categories such as buildings, riders, and cars, especially when compared to predictions at neighboring pixels.

We also investigate the impact of holistic gated feature integration on both Transformer and CNN features. As presented in Table \ref{tb.hgfi}, the baseline setup, which relies solely on the cross-attention mechanism for feature fusion, achieves an mIoU that is 1.75\% lower than the full HGFI strategy. As expected, the holistic gated feature integration of ViT features $\hat{\boldsymbol{F}}_{i+1}^V$ and spatial priors ${\boldsymbol{F}}_{i}^S$ from the side adapter significantly enhances HFIT’s performance.

\section{Conclusion and Future Work}
\label{sec:discussion}
This article discussed a new challenge in computer vision: fully exploiting profound prior knowledge of VFMs for RGB-D driving scene parsing. Our contributions can be summarized in four key aspects: the successful development of an RGB-D scene parsing structure and the introduction of three novel components in the side adapter—DSPE, RHFF, and HGFI. Experimental results demonstrated the significant advantages of HFIT over existing single-modal and data-fusion algorithms, emphasizing its effectiveness and potential for autonomous driving applications.

Looking ahead, advancements in multi-modal VFMs for autonomous driving present new opportunities for robust scene understanding and decision-making. Incorporating emerging techniques like LiDAR-LLM \cite{yang2023lidar} could enable HFIT to simultaneously process LiDAR and depth-based cues, enhancing 3D spatial comprehension in driving environments. Additionally, integrating large language models could improve HFIT's contextual understanding by interpreting ambiguous or complex visual inputs through language-informed perspectives. Furthermore, drawing inspiration from DriveLM’s \cite{sima2023drivelm} graph-based reasoning for sequential decision-making, a promising future direction for HFIT lies in its evolution from a static model to a dynamic system capable of localization, prediction, and action planning. Such advancements would not only increase HFIT's adaptability and responsiveness to real-time challenges but also pave the way for smarter and more interactive autonomous driving systems. Further discussions are provided in the supplementary materials.

\normalem

\bibliographystyle{IEEEtran}
\bibliography{egbib}

\vfill

\end{document}